\documentclass{article}
\usepackage{neurips_2024}
\usepackage[utf8]{inputenc} 
\usepackage[T1]{fontenc}    
\usepackage{url}            
\usepackage{booktabs}       
\usepackage{amsfonts}       
\usepackage{nicefrac}       
\usepackage{microtype}      
\usepackage{xcolor}         
\usepackage{amsmath,amssymb,mathrsfs,subfigure,graphicx,marvosym,algorithm,algorithmic}
\usepackage{footmisc} 
\usepackage{natbib}
\usepackage{algorithmic}
\usepackage{algorithm}

\usepackage{tikz}
\usepackage{changepage}
\usetikzlibrary{positioning}
\setcitestyle{square,numbers}
\title{Demonstration Notebook: Finding the Most Suited In-Context Learning Example from Interactions}

\begin{document}
\bibliographystyle{unsrt}
\maketitle

\begin{abstract}
Large language models (LLMs) benefit greatly from prompt engineering, with in-context learning standing as a pivital technique. While former approaches have provided various ways to construct the demonstrations used for in-context learning, they often ignore the inherent heterogeneity within datasets, applying the same demonstrations to all reasoning questions. We observed that the effectiveness of demonstrations varies depending on the specific question. This motivates our exploration of using prompt engineering to select appropriate demonstrations. To address the challenge of automatically creating and choosing demonstrations tailored to each question, we propose a novel prompt engineering workflow built around a novel object called the "demonstration notebook." This notebook helps identify the most suitable in-context learning example for a question by gathering and reusing information from the LLM's past interactions. Our experiments show that this approach outperforms all existing methods for automatic demonstration construction and selection (as far as we know), achieving state-of-the-art results on serveral reasoning benchmarks. The method's versatility is further demonstrated by its success in text summarization and prompt compression tasks. Additionally, we contribute a rigorous analysis method to reveal the "demonstrative regime" of a demonstration, providing valuable insights into how demonstrations relate to different question types within a dataset.
\end{abstract}

\section{Introduction}
\label{sec:Intro}
The rapidly developing field of large language models (LLMs) has witnessed a surge in the importance of prompt engineering(PE) as different prompt designs demonstrably exert significant influences on LLM outputs. Among the various prompt engineering techniques\cite{Wang2022SelfConsistencyIC,zhou2023leasttomost,chen2022program,wei2023jailbroken}, in-context learning stands as a foundational approach to unlocking LLMs' capabilities for specific tasks requiring the construction and selection of demonstrations.

Introduced in 2020\citep{brown2020language}, in-context learning feeds few-shot examples to language models, enabling them to learn from contextual examples for enhanced performances. To enhance in-context learning capabilities, Chain-of-Thought(ManualCoT)\cite{wei2023chainofthought} prompting advocates to handcraft step-by-step demonstrations within the in-context examples to elicit LLM's stepwise reasoning capabilities for improved performance. Automatic demonstration construction methods have also been proposed. AutoCoT\cite{zhang2022automatic} for example, employes $k$-means clustering to select $k$ demonstrative questions and subsequently generates a $k$-shot demonstration automatically with an LLM. Similarly, PromptSO\cite{Shi2023PromptSO} adopts principle component analysis for the selection of demonstrative questions before automatic demonstration generation. 
While effective in certain scenarios, these approaches rely on fixed demonstrations, neglecting the inherent heterogeneity of questions within a dataset. This limitation hinders further improvement, as is observed different questions benefit from heterogeneous demonstrations. Ideally, the most suitable demonstration is question-specific instead of constant.

Regarding question-specific demonstration selection, retrieval based methods\cite{li2023mot} have garnered significant attention in prompt engineering. These approaches rely on external information retrievers to enhance LLM's output by feeding more information to the model. However, the need of external retrivers necessitates extra efforts to construct a retriever including both the construction of a database and the training of the embeddings which can often be time consuming.
Besides, according to our knowledge, although retrieval based methods have shown effectiveness in various common sense reasoning tasks, little has been explored when it comes to arithmetical and symbolic reasoning tasks since the presence of extra facts might not directly help arithmetic and symbolic reasoning tasks. 
Further more, both retrieval based PE and PE with fixed demonstrations do not directly provide ways to improve model's performances based on problems the LLM fails to solve.

We hypothesize that for arithmetical and symbolic reasoning tasks, demonstrations incorporating proper methodologies, rather than specific factual instances, are more effective. These methodologies can be collected from LLM's former interaction histories and reused as demonstrations with a comprehensive design on the interaction strategy of the LLM. 
To address the challenge of constructing and selecting the most suitable demonstration for a given question, we propose a novel prompt engineering workflow centered around a new object, the \textbf{demonstration notebook}. This notebook encompasses both demonstration generation and selection through three key components: a demonstration set, an interaction record set, and a noted question set. All three sets are initially empty and are automatically populated during a dedicated collection phase preceding the inference stage.

The collection phase of the demonstration notebook consists of several epochs of four procedures each epoch, a demonstration expansion procedure, an on-policy collection procedure, an off-policy collection procedure and a pruning procedure. The demonstration expansion procedure first constructs several one-shot demonstration basis. Several random samples of the basis are concatenated as few-shot demonstrations for the demonstration set. The on-policy and off-policy collection procedures collect more interaction records specifying the questions each demonstration is effective to. After the collection phase, a prompter is trained to select a demonstration from the demonstration set for inference questions. Selected demonstrations will be concatenated with the questions as in-context examples before fed into LLMs in the testing phase.

Our approach is evaluated with several most renowned large language models on a set of nine reasoning benchmarks including: arithmetic reasoning\cite{wang2018translating, roy2015reasoning, cobbe2021training, patel2021nlp, ling2017program}, commonsense reasoning\cite{talmor2019commonsenseqa, geva2021did} and symbolic reasoning\cite{wei2023chainofthought}. Our experimental results show that demonstrations selected by the demonstration notebook consistently outperforms all existing approaches supporting the automatic construction of demonstrations, in align with our hypothesis that heterogeneous demonstration selection can be of vital significance in prompt engineering.

Beyond quantitive experimental results, we also characterize the fact that different demonstrations are effective to heterogeneous sets of questions with a novel concept, \textbf{demonstrative regime}. We provide rigorous experimental results and visualizations to promote more intuitive use of in-context learning examples. Our visualizations show that the demonstrative regimes of demonstrations are often in the form of low dimensional manifolds in the embedding space, which might not be even close to the embeddings of the demonstration itself. This experimental finding holds the potential to revolutionize retrieval and generate(RAG) based prompt engineering methods. For as it indicates, distance in the embedding space might not be a good heuristic for context selection and learning based retrievers should be more suitable compared with more commonly used cosine-similarity based retrievers for the selection of contexts in prompt engineering.

Our contributions are listed as follow:
\begin{itemize} 
\item We introduce a novel prompt engineering method: the demonstration notebook, which tackles the tasks of automatic demonstration construction and selection simultaneously.
\item Through extensive experimentation, we have beaten all the existing methods to the best of our knowledg on demonstration construction which is essentail in prompt engineering.
\item Our work offers the first rigorous analysis and visualization of demonstration regimes for different demonstrations, promoting more intuitive uses in-context examples.
\end{itemize}

\begin{figure}[H]
    \centering
    \includegraphics[width=1\linewidth]{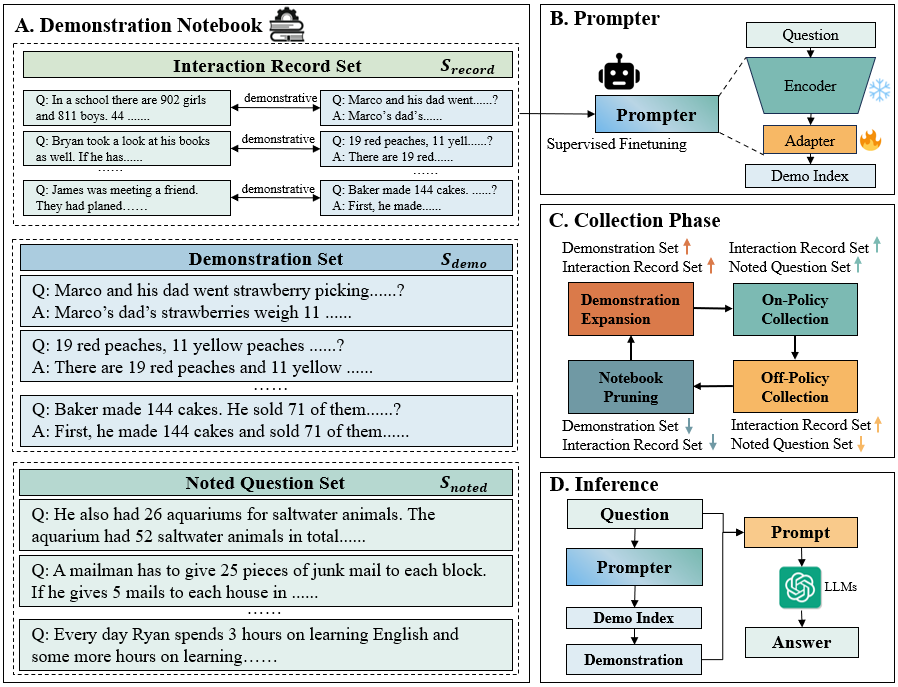}
    \caption{A. The demonstration notebook consists of three components, the interactive record set, the demonstration set and the noted question set. The records in the interactive record set are in the form of (question demonstration) pairs indicating the demonstration is demonstrative to the question. B. The interactive records in the set are used to train a prompter for demonstration selection. We add an adapter module to a pretrained embedding model for this prompter. C. The collection phase of the demonstration notebook consists of several epochs each of four procedures. D. During the inference phase, a question is fed to the prompter for a selected demostration which is subsequently concatenated with the question when fed to LLMs as in-context examples.}
    \label{fig: workflow}
\end{figure}
\section{Related Works}
\label{sec:Related}
The concept of prompt engineering has emerged since 2020\cite{antoun2020aragpt2, muktadir2023brief, brown2020language, logan2021cutting}, as researchers discovered that large language models can learn from demonstrations as contexts. Chain of thought prompting(CoT)\cite{Wei2022ChainOT} suggests hand-crafting demonstrations with nuanced reasoning steps instead of only giving the final answer as demonstrations. Zero-shot CoT\cite{kojima2023large} has shown that nuanced reasoning steps can be generated if provided with proper instructions like "Let's think step by step". In the development, various ensemble prompt engineering methods have been introduced by researchers including CoT-selfconsistency\cite{Wang2022SelfConsistencyIC} and Complexity CoT\cite{fu2023complexitybased}, sampling a diverse set of reasoning paths and voting out a final answer.

Unsupervised learning has been employed for the selection of demonstrations. For instance, Auto-CoT\cite{zhang2022automatic} suggests that we can use k-means clustering to devide the dataset into several clusters and select out representative questions from each clusters to craft demonstrations. PromptSO\cite{Shi2023PromptSO} suggests that we can use principal component analysis in the embedding space and select out the principal components as the questions for demonstration crafting. These methods, though effective, assign the same set of demonstrations to all specific tasks in the dataset and does not take the uniqueness of different questions into consideration when prompting, which might result in lack of performances. Their set of demonstrations are also relatively larger than necessary and thus causes unnecessary token-inefficiency in inference time. 
Reinforcement approaches have also been introduced to train prompters that respond differently based on specificity of questions\cite{deng2022rlprompt, lu2023dynamic}. However they often require a comparatively large training cost to be implemented and can often result in soft prompting instead of grammarly correct prompts.
As shown by several former works, prompt engineering can significantly benefit from extra information collected from former interactions\cite{yao2023tree, besta2023graph, zheng2023progressive}. Their approaches, though effective, requires a long inference time because of requiring multiple interactions with the LLM, limiting their applications to real-world applications like creating AI tutors and AI characters. 

Beyond specific paradigms to select out demonstrations, some general principles about how demonstrations should be formed have been introduced often as certain inductive biases\cite{an2023incontext, wang2023legoprover, goyal2022inductive}, providing heuristics for the process of crafting demonstrations. Instead of heuristic preferences, theoretical analysis of prompt engineering have also been been introduced by researchers. For example, optimal control frameworks have been introduced to systematically analyse prompt engineering\cite{luo2023prompt, bhargava2023whats} and there have also been researchers interpreting in-context learning as implicit gradient decent\cite{dai2023gpt}.

\section{Method}
\label{sec:Method}
This section details our proposed prompt engineering method, the demonstration notebook. We begin with a comprehensive overview in Section~\ref{method:overview}, introducing the demonstration notebook, its components and the four procedures used to collect the notebook, demonstration expansion, on-policy collection, off-policy collection and pruning. Section~\ref{method:expansion} introduces the details for the demonstration expansion procedure. The on-policy and off-policy algorithm for collecting the interaction record set and the noted question set are introduced respectively in Sections~\ref{method:online} and~\ref{method:offline}. Next, Section~\ref{method:prunning} details the pruning procedure used for the demonstration notebook, completing the discussion of the four procedures in the collection phase. Finally, Section~\ref{method:testing} concludes with a description of the prompter and the testing policy. An illustration of the overall workflow is provided in Figure~\ref{fig: workflow}.
\subsection{Overview}
\label{method:overview}
The demonstration notebook is a novel object designed to improve prompt engineering performance through demonstration construction and question-specific selection. The notebook comprises three key components: a demonstration set, an interaction record set, and a noted question set. Additionally, a prompter serves as the demonstration selection policy. All of these three sets contained in the demonstration notebook are collected from the LLM's former interactions with a comprehensive strategy consisted of four procedures iterated for several epochs, including demonstration expansion, on-policy collection, off-policy collection and pruning.

The demonstration set contains multiple demonstrations constructed automatically using the LLM's outputs. Each demonstration is composed of a question and its chain-of-thought answer. This set is initialized and expanded during the demonstration expansion procedure, which occurs at the beginning of each collection epoch.
The interaction record set serves as the supervised training data for the prompter. It accumulates successful problem-solving records from three collection procedures (demonstration expansion, on-policy collection, and off-policy collection). Whenever a demonstration appears to be demonstrative to a question, the record (question, demonstration) will be added to the interaction record set.
The noted question set acts as a repository for "hard" questions encountered during the collection phase. In both the demonstration expansion and the on-policy collection procedures,  multiple demonstrations are tested for a question. If the LLM fails to solve this question using any of the proposed demonstrations, the question is added to the noted question set. Subsequently, questions in the noted question set will be used in off-policy collection as questions to "think twice" and in demonstration expansion as questions for demonstration construction.
\subsection{Demonstration Expansion}
\label{method:expansion}
The demonstration expansion procedure focuses on constructing new demonstrations to enrich the demonstration set.  We begin by sampling several questions that the LLM can correctly solve. These questions serve as a foundation for building demonstrations, each representing a "one-shot" demonstration for the dataset.  Here, we prioritize questions from the noted question set and those residing near the center of question clusters because our empirical observations suggest that these questions are often representative of distinct question clusters and demonstrative to the question clusters they represent. For the noted questions, we prompt the LLM with the correct answer as hints to increase the chances of constructing correct demonstrations for the noted questions. 

Next, we employ random sampling and concatenation to combine several one-shot demonstration bases into a few "few-shot" demonstrations.  These newly constructed demonstrations are then incorporated into the demonstration set of the demonstration notebook.
Following demonstration creation, we establish their demonstrative regimes.  This involves sampling another question set from the dataset's training data and evaluating whether the constructed demonstrations can elicit successful problem-solving of the LLM for these additional questions.
\begin{algorithm}[H]
\caption{Demonstration Expansion}
\begin{algorithmic}[1]
\REQUIRE Trainingset: $S_{train}$; A clustering algorithm: cluster\_center().
\ENSURE $S_{demo}$, $S_{record}$, $S_{noted}$.
\STATE $S_{basis}=S_{noted}\bigcup$ cluster\_centers($S_{train}$).
\STATE \textbf{for} $item$ \textbf{in} $S_{basis}$: \COMMENT{Creating the basis.}
\STATE \quad\indent $ques=item$["$question$"];  $LLManswer=$LLM($ques$).
\STATE \quad\indent \textbf{if} IsCorrect($LLManswer$): $Demo\_basis$.append("$Q:$"+$ques$+"$A:$"+$LLManswer$).
\STATE \textbf{for} $i$ in range NUM\_EXPANSION:  \COMMENT{Creating demonstrations.}
\STATE \quad\indent $ds$ = random.sample($Demo\_basis$,NUM\_PER\_DEMO).
\STATE \quad\indent $demo$ = concatenate($ds$).
\STATE \quad\indent $S_{demo}$.append($demo$).
\STATE \textbf{for} $item$ \textbf{in} $S_{batch}$=sample($S_{train}$):  \COMMENT{Collecting records.}
\STATE \quad\indent $ques=item$["$question$"].
\STATE \quad\indent $Demos=New Demos$.
\STATE \quad\indent $LLManswers=$LLM($Demos$,$ques$).
\STATE \quad\indent \textbf{if} AllCorrect($LLManswers$): pass.
\STATE \quad\indent \textbf{if} PartlyCorrect($LLManswers$): $S_{record}$.append($demo$) if correct.
\STATE \quad\indent \textbf{if} AllWrong($LLManswers$): $S_{noted}.append(ques)$.
\end{algorithmic}
\end{algorithm}

\subsection{On-policy Collection}
\label{method:online}

The on-policy collection procedure targets for the collection of interaction record set of the demonstration notebook in an on-policy way for sample efficiency. We begin by training the prompter based on the current interaction record set which serves as the policy of demonstration selection for both this collection procedure and the testing phase.

In each iteration, a question is sampled from the training set. The prompter then assigns scores to each demonstration in the demonstration set, indicating its predicted suitability for the given question. The top-k demonstrations with the highest scores are then concatenated with the question and presented to the LLM for inference. 


There are three potential outcomes for the inference. First, if all selected demonstrations lead to correct answers, no new records are added to the interaction record set. This scenario suggests that none of the demonstrations are particularly informative for the specific question. Including such interactions could potentially contaminate the prompter's training process, as they don't provide guidance on which demonstration is most effective.

On the other hand, if none of the selected demonstrations enable the LLM to solve the problem, the question is added to the noted question set. These "hard" questions are then prioritized during the off-policy collection procedure, where a wider range of demonstrations can be explored.

The most informative outcome occurs when some, but not all, demonstrations lead to correct answers. This indicates that several demonstrations are demonstrative to the question. Both the question and the successful demonstration(s) are incorporated into the interaction record set. This reinforces the association between demonstrations and their demonstrative regimes, ultimately improving the prompter's selection capability.

\subsection{Off-policy Collection}
\label{method:offline}
The off-policy collection procedure focuses on enriching the demonstration notebook with informative interactions for the noted questions. Unlike the on-policy approach, it does not leverage the prompter for demonstration selection during this stage.

We begin by sampling a batch of questions from the training set, prioritizing questions from the noted question set. Recall that the noted question set contains questions that the LLM struggled with in the on-policy collection procedure. These "hard" questions are prime candidates for exploration with a wider range of demonstrations.

We then systematically evaluate each demonstration in the demonstration set for each question within the batch.  If a demonstration successfully guides the LLM to a correct answer, the question-demonstration pair is added to the interaction record set. This reinforces the association between the demonstrations and their demonstrative regimes and further improves the prompter's selection capability.

\begin{tabular}{p{0.45\textwidth}p{0.45\textwidth}}
\begin{minipage}{.45\textwidth}
\begin{algorithm}[H]
\caption{On-Policy Collection}
\begin{algorithmic}[1]
\REQUIRE Trainingset: $S_{train}$, prompter.
\ENSURE $S_{record}$, $S_{noted}$.
\STATE $S_{batch}=$sample($S_{train}$).
\STATE \textbf{for} $item$ \textbf{in} $S_{batch}$: 
\STATE \quad\indent $ques=item$["$question$"].
\STATE \quad\indent $Demos=$prompter($ques$).
\STATE \quad\indent $LLManswers=$LLM($Demos$,$ques$).
\STATE \quad\indent \textbf{if} AllCorrect($LLManswers$):
\STATE \quad\indent \quad\indent pass.
\STATE \quad\indent \textbf{if} PartlyCorrect($LLManswers$):
\STATE \quad\indent \quad\indent $S_{record}$.append($demo$) if correct.
\STATE \quad\indent \textbf{if} AllWrong($LLManswers$):
\STATE \quad\indent \quad\indent $S_{noted}.append(ques)$.
\end{algorithmic}
\end{algorithm}
\end{minipage}
\end{tabular}
\begin{minipage}{.45\textwidth}
\begin{algorithm}[H]
\caption{Off-Policy Collection}
\begin{algorithmic}[1]
\REQUIRE Trainingset: $S_{train}$, $S_{noted}$, $S_{demo}$.
\ENSURE $S_{record}$, $S_{noted}$.
\STATE $S_{batch}=$sample($S_{noted}$,$S_{train}$).
\STATE \textbf{for} $item$ \textbf{in} $S_{batch}$:
\STATE \quad\indent $ques=item$["$question$"].
\STATE \quad\indent $Demos=S_{demo}$.
\STATE \quad\indent $LLManswers=$LLM($Demos$,$ques$).
\STATE \quad\indent \textbf{if} AnyCorrect($LLManswers$):
\STATE \quad\indent \quad\indent $S_{noted}.pop(ques)$.
\STATE \quad\indent \textbf{if} PartlyCorrect($LLManswers$):
\STATE \quad\indent \quad\indent $S_{record}$.append($demo$) if correct.
\STATE \quad\indent \textbf{if} AllWrong($LLManswers$):
\STATE \quad\indent \quad\indent $S_{noted}.append(ques)$.
\end{algorithmic}
\end{algorithm}
\end{minipage}

\subsection{Pruning}
\label{method:prunning}
Each epoch of the collection phase concludes with a pruning procedure to eliminate redundant records and demonstrations within the demonstration notebook.

Interaction Record Set: For the interaction record set, we identify and remove any duplicate records.  These duplicates may arise due to coinciding samples during previous collection procedures. This ensures the interaction record set contains unique question-demonstration pairs for the prompter.

Demonstration Set: For each demonstration in the set, we define a corresponding "coverage set" consisting of all questions that the LLM can solve correctly when aided by that demonstration. During pruning, if the coverage set of one demonstration is entirely contained within the coverage set of another demonstration, the redundant demonstration is removed from the demonstration set. Additionally, all associated records from the interaction record set are cleared.

\begin{algorithm}[H]
\caption{Pruning}
\begin{algorithmic}[1]
\REQUIRE Trainingset: $S_{record}$, $S_{noted}$, $S_{demo}$.
\ENSURE $S_{record}$, $S_{noted}$, $S_{demo}$.
\STATE \textbf{for} $record$ \textbf{in} $S_{record}$:  \COMMENT{Pruning redundant records.}
\STATE \quad\indent \textbf{if} IsDuplicated($record$): $S_{record}$.pop($record$).
\STATE \textbf{for} $demo$ \textbf{in} $S_{demo}$:  \COMMENT{Calculating demonstrative regimes.}
\STATE \quad\indent $S_{regime}$["$demo$"] = {$rec$["$question$"] for $rec$ in $S_{record}$ with $rec$["$demonstration$"]==$demo$}.
\STATE \textbf{for} $demo_1$, $demo_2$ \textbf{in} $S_{demo}$: 
\STATE \quad\indent \textbf{if} $S_{regime}$["$demo_1$"] included in $S_{regime}$["$demo_2$"]: \COMMENT{Pruning redundant demonstrations.}
\STATE \quad\indent \quad\indent $S_{demo}$.pop($demo_1$).
\STATE \quad\indent \quad\indent $S_{rec}$.pop({$rec$ for $rec$ in $S_{record}$ with $rec$["$demonstration$"]==$demo_1$}).
\end{algorithmic}
\end{algorithm}

\subsection{Training the Prompter and Inference}
\label{method:testing}
Following the collection phase, we train the prompter of the demonstration notebook using the interaction record set. This empowers the prompter to automatically select the most suitable demonstration for a given question during inference.

To balance sample efficiency and computational training costs, we propose a prompter architecture that leverages a pre-trained text encoder with several shallow adapter layers. The interaction record set, while valuable, is relatively small for comprehensive training of a large prompter, particularly one that aims to capture the full complexity of prompts. Therefore, utilizing a pre-trained encoder alongside a smaller adapter appears to be operable and efficient.

During the testing phase, when encountering a new question, the prompter infers and selects the most suited demonstration from the demonstration set. This chosen demonstration is then concatenated with the question to form the final prompt that is fed to the LLM for inference.

\section{Experiments}
\label{sec:Experiment}
We evaluate the effectiveness of demonstration notebook on several settings including three types of reasoning tasks, prompt compression and article sumarization. As former works pointed out, language models often struggle at reasoning tasks including arithmetic reasoning, commonsense reasoning, and symbolic reasoning, making reasoning tasks a common playground to testify LLM's performances. Recent research has suggested that since the use of large language models can be costly in computation and resources, prompt compression, aiming at compressing the length of the prompts used in inference while maintaining good performances, can be of vital importance. We also extend our experimental results to article summarization aiming at concise summarization of texts in natural language.
\subsection{In-context Learning for Reasoning}
The demonstration notebook is evaluated on various reasoning tasks against relevant baselines that can also achieve automatic demonstration construction.

\textbf{Benchmarks} To evaluate the effectiveness of demonstration notebook in reasoning tasks, we use the following benchmarks consisted of three main categories:

1. Arithmetic reasoning: (1) SingleEq\cite{wang2018translating}, (2) MultiArith\cite{roy2015reasoning}, (3) GSM8k\cite{cobbe2021training}, (4) SVAMP\cite{patel2021nlp}, (5) AQuA\cite{ling2017program}. These datasets each consists of arithmetic questions of secondary school math difficulty. 

2. Commonsense reasoning: (1) CSQA\cite{talmor2019commonsenseqa}, (2) STQA\cite{geva2021did}. The popular CSQA\cite{talmor2019commonsenseqa} contains commonsense questions about the world involving complex semantics that often require prior knowledge in order to answer correcly. StrategyQA\cite{geva2021did} requires models to infer a multi-step reasoning path to answer questions by stating out the possibility of some statements.

3. Symbolic reasoning: (1) Last Letter Concatencation and (2) Coin Flip\cite{wei2023chainofthought}. The last letter concatenation task asks the LLM to concatenate the last letters of each words in a word list which often requires step by step reasoning and can benefit significantly from in-context examples. The coin flip task inquires the LLM to make predictions about whether a coin is facing head up or head down after multiple operations.

\textbf{Baselines} Our experimental results are compared with four baselines using in-context learning to enhance LLM performances. First, Zero-shot CoT\cite{kojima2023large} prompts the LLM with the question and a specially designed instruction, "Let's think step by step." Manual CoT\cite{wei2023chainofthought} uses chain-of-thought demonstrations crafted with extra human expertise. AutoCoT\cite{zhang2022automatic} uses k-means clustering to select out several questions and uses the LLM to automatically construct the demonstrations while PromptSO\cite{Shi2023PromptSO} uses principle component analysis for the selection of questions. For ManuelCoT\cite{wei2023chainofthought}, AutoCoT\cite{zhang2022automatic} and PromptSO\cite{Shi2023PromptSO}, we use exactly the same demonstrations provided in their works respectively.

\textbf{Implement Details}
We evaluate our method on Meta Llama3(8B) model and OpenAI gpt-3.5-turbo(175B) model. As former works pointed out, they each stand as representative large language models of open-source and close-source communities.
We use greedy decoding in inference for deterministic text generation and reproductivity.
The prompter is tuned based on OpenAI's text-embedding-ada-002 model with an MLP adapter at the head of the embedding model.

\subsubsection{Main Results}
\label{llmreasoning}

\begin{table}[H]
    \caption{Main Experimental Results with Llama3-8B}
    \label{tab: Data}
    \centering
    \begin{tabular}{c c c c c c c c c c c}
    \toprule
     \bf{Method} &SingleEq&MultiArith&GSM8k&SVAMP&AQuA& CSQA & STQA & Letter & Avg.\\
     \midrule
     ZeroCoT     & 92.7\% &98.8\%& 72.3\% & 79.8\%& 50.0\%& 53.4\%& 56.7\%& 47.3\%& 65.71\%\\
     ManualCoT   & 93.3\% &98.3\%& 73.3\% & 81.8\%& 46.9\%& 66.4\%& 67.2\%& 75.3\%& 75.31\%\\
     AutoCoT     & 96.8\% &98.8\%& 73.4\% & 82.4\%& 47.8\%& 72.7\%& 71.6\%& 76.7\%& 76.30\%\\
     PromptSO    & 94.2\% &99.4\%& 77.6\% & 83.0\%& 52.0\%& 57.6\%& 69.4\%& 81.3\%& 75.44\%\\
     \midrule
     Ours        & 97.2\% &99.4\%& 74.8\% & 84.2\%& 47.6\%& 68.7\%& 70.7\%& 94.7\%& 79.66\%\\
     \bottomrule
    \end{tabular}
\end{table}

\begin{table}[H]
    \caption{Main Experimental Results with GPT3.5}
    \label{tab: Data}
    \centering
    \begin{tabular}{c c c c c c c c c c c}
    \toprule
     \bf{Method}  & SingleEq & MultiArith& GSM8k & SVAMP & AQuA  & CSQA  & STQA  &  Letter & Avg.\\ 
     \midrule
     ZeroCoT      & 90.4\%   & 96.0\%    &75.1\% & 76.5\%& 38.2\%& 72.0\%& 57.6\%& 71.0\%& 72.10\%\\
     ManualCoT    & 90.9\%   & 97.0\%    &75.8\% & 80.2\%& 45.3\%& 72.2\%& 61.7\%& 78.8\%& 75.24\%\\
     AutoCoT      & 90.2\%   & 96.0\%    &74.1\% & 78.2\%& 46.5\%& 72.7\%& 62.4\%& 72.6\%& 74.09\%\\
     PromptSO     & 92.1\%   & 98.8\%    &77.9\% & 81.0\%& 40.6\%& 74.1\%& 62.5\%& 79.6\%& 75.83\%\\
     \midrule
     Ours         & 98.8\%   & 99.4\%    &70.8\% & 80.4\%& 60.2\%& 70.6\%& 71.2\%& 93.3\%& 80.59\%\\
     \bottomrule
    \end{tabular}
\end{table}

\subsection{In-context Learning for Article Summarization}
The demonstration notebook can also be applied to article summarization tasks aiming at concise summarization of texts in natural language. We use Samsung Abstractive Messenger Summarization\cite{gliwa-etal-2019-samsum} as the testbed and compare demonstrations seleted by the collected demonstration notebook with randomly chosen demonstrations and zero-shot prompting. We use the ROUGE metric for the evaluation of summarization quality.

In this setting, since the ROUGE score is calculated to be ranging from 0 to 1. We slightly adjust the collection criteria of the demonstration notebook. If a summarization is above a predefined number, we regard it as corret in the process.

Our experimental results show that with minor modifications of the algorithm, the demonstration notebook can be applied to situations with continuous evaluations. 
\begin{table}[H]
    \caption{Article Summarization ROUGEs with Llama3-8B}
    \label{tab: Data}
    \centering
    \begin{tabular}{c c c}
    \toprule
     \bf{Method}            & SAMsun&   \\
     \midrule
     Zero-shot              & 32.8  &   \\
     Random Demonstration   & 33.4  &   \\
     Demonstration Notebook & 35.8  &   \\
     \bottomrule
    \end{tabular}
\end{table}
\subsection{Further Analysis on the Demonstrative Regime}
In order to characterize the phenomenon that different demonstrations are effective to different kinds of questions, we propose to use the concept of demonstrative regime. By definition, if a question $q$ which can not be directly solved by an LLM can be solved by the same LLM with the presence of a demonstration $d$, we say that $d$ is \textbf{demonstrative} to $q$ for this LLM. All the questions that the demonstration $d$ is demonstrative to form a set noted as demonstrative regime of demonstration $d$.
Notice that if the question $q$ is simple enough, a demonstration may not be needed for an LLM to answer correctly. This is the main reason why we eliminate these questions that can be easily solved by an LLM when considering the demonstrative regimes. With this perspective, the core idea of demonstration notebook can be viewed as to construct a complete demonstration set so that the union of their demonstrative regimes can cover the whole distribution of the dataset, and the interaction record set is aimed to specify the demonstrative regimes from interactions by collecting points from each demonstrative regime.

\begin{figure}[H]
    \centering
    \includegraphics[width=1\linewidth]{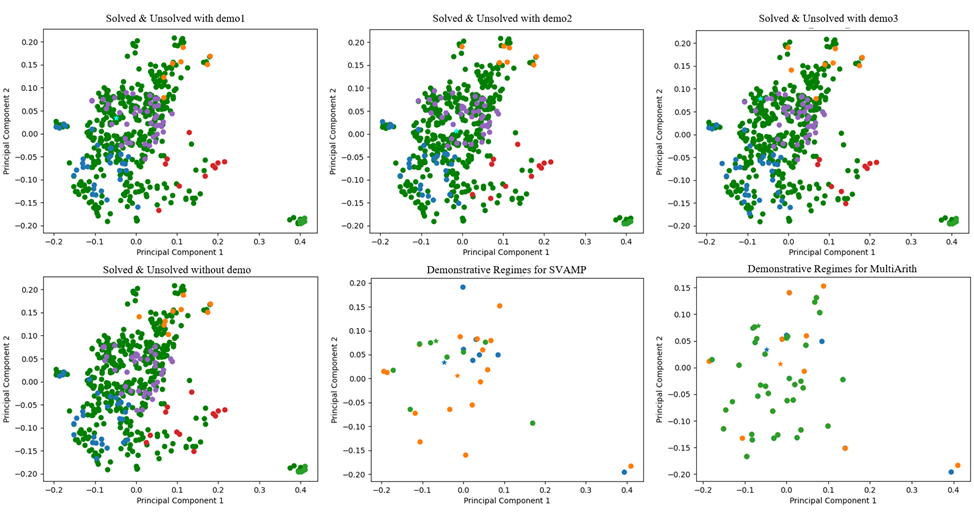}
    \caption{Visualization of questions that can be solved by the LLM directly, and using three different 
demonstrations respectively. The green dots means the question that can be correctly solved and ther
other dots refer to clusters of questions that can be solved with the demonstrations. The last two 
diagrams refer to the demonstrative regiems of three demonstrations for two datasets, SVAMP and 
MutlitArith.}
    \label{fig:regime}
\end{figure}

We also provide visualizations of the demonstrative regimes for more intuitive understanding of the concept of demonstrative regimes\ref{fig:regime}. By transforming the demonstrations into their embeddings and visualizing the first two principal components, we see several characteristics of a demonstrative regime:
1. The demonstrative regime of a demonstration forms a relatively small area in the embedding space and universal demonstrative regimes are uncommon. 
2. The demonstrative regime of a demonstration might not be close to the demonstration in the embedding space.
3. The demonstrative regime of a demonstration is likely to be a low dimensional manifold containing this demonstration itself.


\section{Conclusion}
In this work, we propose a novel method, demonstration notebook to tackle automatic demonstration generation and selection using information collected from former interactions. Our experimental results have achieved SOTA result in demonstration construction beating all other approaches that support automatic demonstration generation. The demonstration notebook algorithm can be extended to other tasks including prompt compression and article sumarization and our experimental results have shown the effectiveness of this appraoch in these settings proving the versatility of the method. We also pioneer in exploring the concept of demonstrative regimes characterizing the questions that can be solved with the presence of a demonstration. Our visualization results provide valuable insights toward the intuitive understanding of using demonstrations in in-context learning.

\newpage
\bibliography{main}
\end{document}


\bibliographystyle{unsrt}

\section{Appendix Related Works}
\section{Theoretical Analysis of Demonstrative Regimes}
\section{Appendix Visualizations}
\section{Appendix Experimental Results}

\newpage
\bibliography{ref}